\documentclass[11pt]{article}
\usepackage{eacl2017}
\usepackage{times}
\usepackage{url}
\usepackage{latexsym}
\usepackage{relsize}

\usepackage{multirow}
\usepackage{tikz}
\usetikzlibrary{arrows,shapes,trees}
\usepackage{fixltx2e}
\usepackage{amsmath}
\usepackage{amssymb}
\usepackage{enumitem}
\usepackage{bold-extra}
\usepackage{rotating}
\usepackage{tikz-dependency}

\DeclareMathOperator*{\argmax}{arg\,max}

\eaclfinalcopy 

\title{Dependency Parsing as Head Selection}

\author{Xingxing Zhang, Jianpeng Cheng \and Mirella Lapata \\
	Institute for Language, Cognition and Computation \\
	School of Informatics, University of Edinburgh \\ 
	10 Crichton Street, Edinburgh EH8 9AB \\
	{\tt \{x.zhang,jianpeng.cheng\}@ed.ac.uk, mlap@inf.ed.ac.uk} }

\date{}

\begin{document}
\maketitle
\begin{abstract}
  Conventional graph-based dependency parsers guarantee a tree
  structure both during training and inference. Instead, we formalize
  dependency parsing as the problem of independently selecting the
  head of each word in a sentence. Our model which we call
  \textsc{DeNSe} (as shorthand for {\bf De}pendency {\bf N}eural {\bf
    Se}lection) produces a distribution over possible heads for each
  word using features obtained from a bidirectional recurrent neural
  network. Without enforcing structural constraints during training,
  \textsc{DeNSe} generates (at inference time) trees for the
  overwhelming majority of sentences, while non-tree outputs can be
  adjusted with a maximum spanning tree algorithm.  We evaluate
  \textsc{DeNSe} on four languages (English, Chinese, Czech, and
  German) with varying degrees of non-projectivity. Despite the
  simplicity of the approach, our parsers are on par with the state of
  the art.\footnote{Our code is available at
    \url{http://github.com/XingxingZhang/dense_parser}.}
\end{abstract}

\section{Introduction}
Dependency parsing plays an important role in many natural language
applications, such as relation extraction \cite{fundel2007relex},
machine translation \cite{carreras2009non}, language modeling \cite{chelba1997structure,zhang-etal:2016} and ontology construction
\cite{snow2004learning}.  Dependency parsers represent syntactic
information as a set of head-dependent relational arcs, typically
constrained to form a tree.  Practically all models proposed for
dependency parsing in recent years can be described as graph-based
\cite{mcdonald2005online} or transition-based
\cite{yamada2003statistical,nivre2006labeled}.

Graph-based dependency parsers are typically arc-factored, where the
score of a tree is defined as the sum of the scores of all its
arcs. An arc is scored with a set of local features and a linear
model, the parameters of which can be effectively learned with online
algorithms
\cite{crammer2001algorithmic,crammer2003ultraconservative,freund1999large,collins2002discriminative}. In
order to efficiently find the best scoring tree during training
\emph{and} decoding, various maximization algorithms have been
developed
\cite{eisner1996three,eisner2000bilexical,mcdonald2005non}. In
general, graph-based methods are optimized globally, using features of
single arcs in order to make the learning and inference tractable.
Transition-based algorithms factorize a tree into a set of parsing
actions. At each transition state, the parser scores a candidate
action conditioned on the state of the transition system and the
parsing history, and greedily selects the highest-scoring action to
execute. This score is typically obtained with a classifier based on
non-local features defined over a rich history of parsing decisions
\cite{yamada2003statistical,zhang2011transition}.


Regardless of the algorithm used, most well-known dependency parsers,
such as the \mbox{MST-Parser} \cite{mcdonald2005non} and the MaltPaser
\cite{nivre2006maltparser}, rely on extensive feature engineering.
Feature templates are typically manually designed and aim at capturing
head-dependent relationships which are notoriously sparse and
difficult to estimate. More recently, a few approaches
\cite{chen2014fast,pei2015effective,DBLP:journals/corr/KiperwasserG16a}
apply neural networks for learning dense feature representations. The
learned features are subsequently used in a conventional graph- or
transition-based parser, or better designed variants
\cite{dyer2015transition}.


In this work, we propose a simple neural network-based model which
learns to select the head for each word in a sentence without
enforcing tree structured output.
Our model which we call {\sc DeNSe} (as shorthand for {\bf De}pendency
{\bf N}eural {\bf Se}lection) employs bidirectional recurrent neural
networks to learn feature representations for words in a
sentence. These features are subsequently used to predict the head of
each word.
Although there is nothing inherent in the model to enforce
tree-structured output, when tested on an English dataset, it is able
to generate trees for~95\% of the sentences, 87\% of which are
projective. The remaining non-tree (or non-projective) outputs are
post-processed with the Chu-Liu-Edmond (or Eisner) algorithm.  {\sc
  DeNSe} uses the head selection procedure to estimate arc weights
during training. During testing, it essentially reduces to a standard
graph-based parser when it fails to produce tree (or projective)
output.



We evaluate our model on benchmark dependency parsing corpora,
representing four languages (English, Chinese, Czech, and German) with
varying degrees of non-projectivity.  Despite the simplicity of our
approach, experiments show that the resulting parsers are on par with the
state of the art.
%
%
%
%
%

\section{Related Work}


\paragraph{Graph-based Parsing} Graph-based dependency parsers employ
a model for scoring possible dependency graphs for a given
sentence. The graphs are typically factored into their component arcs
and the score of a tree is defined as the sum of its arcs.  This
factorization enables tractable search for the highest scoring graph
structure which is commonly formulated as the search for the maximum
spanning tree (MST).
The Chu-Liu-Edmonds algorithm
\cite{chu1965shortest,edmonds1967optimum,mcdonald2005non} is often
used to extract the MST in the case of non-projective trees, and the
Eisner algorithm \cite{eisner1996three,eisner2000bilexical} in the
case of projective trees.  During training, weight parameters of the
scoring function can be learned with margin-based algorithms
\cite{crammer2001algorithmic,crammer2003ultraconservative} or the
structured perceptron
\cite{freund1999large,collins2002discriminative}. Beyond basic
first-order models, the literature offers a few examples of
higher-order models involving sibling and grand parent relations
\cite{carreras2007experiments,koo2010dual,zhang2012generalized}.
Although more expressive, such models render both training and
inference more challenging.




\paragraph{Transition-based Parsing}
As the term implies, transition-based parsers conceptualize the
process of transforming a sentence into a dependency tree as a
sequence of transitions. A transition system typically includes a
stack for storing partially processed tokens, a buffer containing the
remaining input, and a set of arcs containing all dependencies between
tokens that have been added so far
\cite{nivre2003efficient,nivre2006labeled}.  A dependency tree is
constructed by manipulating the stack and buffer, and appending arcs
with predetermined operations. Most popular parsers employ an
\textit{arc-standard}
\cite{yamada2003statistical,nivre2004incrementality} or
\textit{arc-eager} transition system \cite{nivre2008algorithms}.
Extensions of the latter include the use of non-local training methods
to avoid greedy error propagation
\cite{zhang2008tale,huang2010dynamic,zhang2011transition,goldberg2012dynamic}.

\paragraph{Neural Network-based Features} 
Neural network
representations have a long history in syntactic parsing
\cite{Mayberry:Miikkulainen:1999,henderson:2004:ACL,titov-henderson:2007:ACLMain}.
Recent work uses neural networks in lieu of the linear classifiers
typically employed in conventional transition- or graph-based
dependency parsers.
For example, \newcite{chen2014fast} use a feed forward neural network
to learn features for a transition-based parser, whereas
\newcite{pei2015effective} do the same for a graph-based
parser. \newcite{lei2014low} apply tensor decomposition to obtain word
embeddings in their syntactic roles, which they subsequently use in a
graph-based parser. \newcite{dyer2015transition} redesign components
of a transition-based system where the buffer, stack, and action
sequences are modeled separately with stack long short-term memory
networks. The hidden states of these LSTMs are concatenated and used
as features to a final transition
classifier. \newcite{DBLP:journals/corr/KiperwasserG16a} use
bidirectional LSTMs to extract features for a transition- and
graph-based parser, whereas \newcite{cross-huang:2016} build a greedy
arc-standard parser using similar features.


In our work, we formalize dependency parsing as the task of finding
for each word in a sentence its most probable head. Both head
selection and the features it is based on are learned using neural
networks. The idea of modeling child-parent relations independently dates back to
\newcite{hall:2007} who use an edge-factored model to generate
$k$-best parse trees which are subsequently reranked using a model
based on rich global features. Later \newcite{smith2010efficient} show
that a head selection variant of their loopy belief propagation parser
performs worse than a model which incorporates tree structure
constraints. Our parser is conceptually simpler: we rely on head
selection to do most of the work and decode the best tree
\emph{directly} without using a reranker. In common with recent neural network-based dependency
parsers, we aim to alleviate the need for hand-crafting feature
combinations. Beyond feature learning, we further show that it is
possible to simplify the training of a graph-based dependency parser
in the context of bidirectional recurrent neural networks.




\section{Dependency Parsing as Head Selection}


In this section we present our parsing model, {\sc DeNSe}, which tries
to predict the head of each word in a sentence. Specifically, the
model takes as input a sentence of length~$N$ and outputs~$N$
$\langle$head, dependent$\rangle$ arcs. We describe the model focusing
on unlabeled dependencies and then discuss how it can be
straightforwardly extended to the labeled setting.  We begin by
explaining how words are represented in our model and then give
details on how {\sc DeNSe} makes predictions based on these learned
representations. Since there is no guarantee that the outputs of {\sc
	DeNSe} are trees (although they mostly are), we also discuss how to
extend {\sc DeNSe} in order to enforce projective and non-projective
tree outputs.  Throughout this paper, lowercase boldface letters
denote vectors (e.g.,~$\mathbf{v}$ or $\mathbf{v}_i$), uppercase
boldface letters denote matrices (e.g.,~$\mathbf{M}$ or
$\mathbf{M}_b$), and lowercase letters denote scalars (e.g.,~$w$ or
$w_i$).


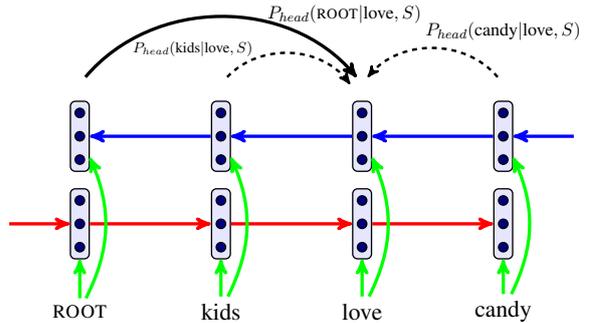
\begin{figure}
	\centering
	
	\resizebox{0.48\textwidth}{!}{%
		
		\centering
		\begin{tikzpicture}[scale=.75,->,>=stealth',thick,main node/.style={rectangle,rounded corners=3pt,fill=blue!10,draw,font=\sffamily\Large\bfseries,inner sep=0,minimum size=2.5mm,minimum width=4mm,minimum height=1.5cm,path picture={
				\draw[fill=blue!50!black] (0, -0.5) circle (1mm);
				\draw[fill=blue!50!black] (0, 0) circle (1mm);
				\draw[fill=blue!50!black] (0, 0.5) circle (1mm);
			}
		}]
		
		\node[main node] (h0) at (-6, 0) {};
		\node[main node] (h1) at (-2, 0) {};
		\node[main node] (h2) at (2, 0) {};
		\node[main node] (h3) at (6, 0) {};
		
		\node[main node] (h0_) at (-6, 2.5) {};
		\node[main node] (h1_) at (-2, 2.5) {};
		\node[main node] (h2_) at (2, 2.5) {};
		\node[main node] (h3_) at (6, 2.5) {};
		
		\node (w0) at (-6, -2.5) {\Large {\sc root}};
		\node (w1) at (-2, -2.5) {\Large kids};
		\node (w2) at (2, -2.5) {\Large love};
		\node (w3) at (6, -2.5) {\Large candy};
		
		\draw[line width=2pt,green] (w0) -- (h0);
		\draw[line width=2pt,green] (w1) -- (h1);
		\draw[line width=2pt,green] (w2) -- (h2);
		\draw[line width=2pt,green] (w3) -- (h3);
		
		\draw[line width=2pt,red] (-8, 0) -- (h0);
		\draw[line width=2pt,red] (h0) -- (h1);
		\draw[line width=2pt,red] (h1) -- (h2);
		\draw[line width=2pt,red] (h2) -- (h3);
		
		\draw[line width=2pt,blue] (8, 2.5) -- (h3_);
		\draw[line width=2pt,blue] (h3_) -- (h2_);
		\draw[line width=2pt,blue] (h2_) -- (h1_);
		\draw[line width=2pt,blue] (h1_) -- (h0_);
		
		\path (w0) edge[bend right=25,line width=2pt,green] (h0_);
		\path (w1) edge[bend right=25,line width=2pt,green] (h1_);
		\path (w2) edge[bend right=25,line width=2pt,green] (h2_);
		\path (w3) edge[bend right=25,line width=2pt,green] (h3_);
		
		\node (h0_fb) at (-6, 4) {};
		\node (h1_fb) at (-2, 4) {};
		\node (h2_fb) at (2, 4) {};
		\node (h2_fb_) at (1.8, 3.8) {};
		\node (h3_fb) at (6, 4) {};
		
		\path (h0_fb) edge[bend left=50,line width=2pt] (h2_fb);
		\path (h1_fb)[dashed] edge[bend left=50,line width=1.5pt] (h2_fb_);
		\path (h3_fb)[dashed] edge[bend right=50,line width=1.5pt] (h2_fb);
		
		\node (p_0_2) at (1.5, 6) { $P_{head}(\text{\sc root}|\text{love}, S)$};
		
		\node (p_1_2) at (-2.8, 5) {\small $P_{head}(\text{kids}|\text{love}, S)$};
		
		\node (p_3_2) at (6, 5.5) { $P_{head}(\text{candy}|\text{love}, S)$};
		
		\end{tikzpicture}
		
	}%
	
	\caption{{\sc DeNSe} estimates the probability a word being
          the head of another word based on bidirectional LSTM
          representations for the two words.  $P_{head}(\text{\sc
            root}|\text{love}, S)$ is the probability of {\sc root}
          being the head of \textit{love} (dotted arcs denote
          candidate heads; the solid arc is the goldstandard).}
	\label{fig:dense}
\end{figure}

\subsection{Word Representation}
\label{sec:wordrepr}
Let $S=(w_0, w_1, \dots, w_N)$ denote a sentence of length~$N$;
following common practice in the dependency parsing literature
\cite{Kubler:etal:2009}, we add an artificial {\sc root} token
represented by~$w_0$. Analogously, let \mbox{$A=(\mathbf{a}_0,
  \mathbf{a}_1, \dots, \mathbf{a}_N)$} denote the representation of
sentence~$S$, with $\mathbf{a}_i$ representing word~$w_i \quad (0 \le
i \le N)$. Besides encoding information about each $w_i$~in isolation
(e.g.,~its lexical meaning or POS tag), $\mathbf{a}_i$ must also
encode~$w_i$'s positional information within the sentence. Such
information has been shown to be important in dependency parsing
\cite{mcdonald2005online}. For example, in the following sentence:

\begin{center}
	\begin{dependency}[theme = simple, arc angle=50]
		\begin{deptext}[column sep=0.03em]
			{\sc root} \&  a \& dog \& is \& chasing \& a \& cat \\
		\end{deptext}
		\depedge{3}{2}{}
		\depedge{5}{3}{}
		\depedge{5}{4}{}
		\depedge{1}{5}{}
		\depedge{7}{6}{}
		\depedge{5}{7}{}
	\end{dependency}
\end{center}
%
%
the head of the first \textsl{a} is \textsl{dog}, whereas the head of
the second \textsl{a} is \textsl{cat}.  Without considering positional
information, a model cannot easily decide which \textsl{a} (nearer or
farther) to assign to \textsl{dog}.

Long short-term memory networks (Hochreiter and Schmidhuber, 1997;
LSTMs), a type of recurrent neural network with a more complex
computational unit, have proven effective at capturing long-term
dependencies.  In our case LSTMs allow to represent each word on its
own and within a sequence leveraging long-range contextual
information.  As shown in Figure~\ref{fig:dense}, we first use a
forward LSTM ($\text{LSTM}^F$) to read the sentence from left to right
and then a backward LSTM ($\text{LSTM}^B$) to read the sentence from
right to left, so that the entire sentence serves as context for each
word:\footnote{For more detail on LSTM networks, see e.g.,
  \newcite{Graves:2012} or \newcite{Goldberg:2015}.}
\begin{equation}
	\label{eq:forward}
	\mathbf{h}_i^F,  \mathbf{c}_i^F = \text{LSTM}^F(\mathbf{x}_i, \mathbf{h}_{i-1}^F, \mathbf{c}_{i-1}^F)
\end{equation}
\begin{equation}
	\label{eq:backward}
	\mathbf{h}_i^B,  \mathbf{c}_i^B = \text{LSTM}^B(\mathbf{x}_i, \mathbf{h}_{i+1}^B, \mathbf{c}_{i+1}^B)
\end{equation}
where $\mathbf{x}_i$ is the feature vector of word $w_i$, $\mathbf{h}_i^F \in \mathbb{R}^{d}$ and
$\mathbf{c}_i^F \in \mathbb{R}^{d}$ are the hidden states and memory
cells for the $i$th word $w_i$ in $ \text{LSTM}^F$ and~$d$ is the
hidden unit size. $\mathbf{h}_i^F$ is also the representation for
$w_{0:i}$ ($w_i$ and its left neighboring words) and $\mathbf{c}_i^F$
is an internal state maintained by $ \text{LSTM}^F$.
$\mathbf{h}_i^B \in \mathbb{R}^d$ and
$\mathbf{c}_i^B \in \mathbb{R}^d$ are the hidden states and memory
cells for the backward $\text{LSTM}^B$.  Each token~$w_i$ is
 represented by $\mathbf{x}_i$, the concatenation of two
vectors corresponding to $w_i$'s lexical and POS tag embeddings:
\begin{equation}
	\label{eq:fearep}
	\mathbf{x}_i = [ \mathbf{W}_e \cdot e(w_i); \mathbf{W}_t \cdot e(t_i) ]
\end{equation}
where $e(w_i)$ and $e(t_i)$ are one-hot vector representations
of token~$w_i$ and its POS tag $t_i$;
$\mathbf{W}_e \in \mathbb{R}^{s \times |V|}$ and
$\mathbf{W}_t \in \mathbb{R}^{q \times |T|}$ are the word and POS tag
embedding matrices, where $|V|$ is the vocabulary size, $s$ is the
word embedding size, $|T|$ is the POS tag set size, and $q$ the tag
embedding size. The hidden states of the forward and backward LSTMs
are concatenated to obtain$~\mathbf{a}_i$, the final representation of
$w_i$:
\begin{equation}
	\label{eq:ann}
	\mathbf{a}_i = [ \mathbf{h}_i^F; \mathbf{h}_i^B ] \quad i \in [0, N]
\end{equation}

Note that bidirectional LSTMs are one of many possible ways of
representing word~$w_i$. Alternative representations include
embeddings obtained from feed-forward neural networks
\cite{chen2014fast,pei2015effective}, character-based embeddings
\cite{ballesteros-dyer-smith:2015:EMNLP}, and more conventional
features such as those introduced in \newcite{mcdonald2005online}.

\subsection{Head Selection}

%

We now move on to discuss our formalization of dependency parsing as
head selection. We begin with unlabeled dependencies and then explain
how the model can be extended to predict labeled ones.

In a dependency tree, a head can have multiple
dependents, whereas a dependent can have only one head. Based on this
fact, dependency parsing can be formalized as follows. Given a
sentence $S=(w_0, w_1, \dots, w_N)$, we aim to find for each word $w_i
\in \{w_1, w_2, \dots, w_n\}$ the most probable head $w_j \in \{w_0,
w_1, \dots, w_N\}$. For example, in Figure~\ref{fig:dense}, to find
the head for the token \textit{love}, we calculate probabilities
$P_{head}(\text{\sc root}|\text{love}, S)$,
$P_{head}(\text{kids}|\text{love}, S)$, and
$P_{head}(\text{candy}|\text{love}, S)$, and select the highest.  More
formally, we estimate the probability of token~$w_j$ being the head
of token~$w_i$ in sentence~$S$ as:
\begin{equation}
	\label{eq:softmax}
	P_{head}(w_j|w_i, S) = \frac{\exp(g(\mathbf{a}_j, \mathbf{a}_i))}{ \sum_{k=0}^{N} \exp( g(\mathbf{a}_k, \mathbf{a}_i) ) }
\end{equation}
where $\mathbf{a}_i$ and $\mathbf{a}_j$ are vector-based
representations of $w_i$ and $w_j$, respectively (described in
Section~\ref{sec:wordrepr}); $g(\mathbf{a}_j, \mathbf{a}_i)$ is a
neural network with a single hidden layer that computes the
associative score between representations~$\mathbf{a}_i$ and
$\mathbf{a}_j$:
\begin{equation}
	\label{eq:tanh}
	g(\mathbf{a}_j, \mathbf{a}_i) = \mathbf{v}_a^\top \cdot \tanh( \mathbf{U}_a \cdot \mathbf{a}_j + \mathbf{W}_a \cdot \mathbf{a}_i )
\end{equation}
where $\mathbf{v}_a \in \mathbb{R}^{2d}$, $\mathbf{U}_a \in
\mathbb{R}^{2d \times 2d}$, and $\mathbf{W}_a \in \mathbb{R}^{2d \times
	2d}$ are weight matrices of $g$.  Note that the candidate head $w_j$
can be the {\sc root}, while the dependent~$w_i$ cannot.
Equations~\eqref{eq:softmax} and~\eqref{eq:tanh} compute
the probability of adding an arc between two words, in a fashion
similar to the neural attention mechanism in sequence-to-sequence
models \cite{bahdanau:2014}.


We train our model by minimizing the negative log likelihood of the
gold standard $\langle$head, dependent$\rangle$ arcs in all training
sentences:
\begin{equation}
\hspace*{-2ex}J(\theta) = - \frac{ 1 }{  |\mathcal{T}| } \sum_{S \in \mathcal{T}} \sum_{i=1}^{N_S} \log P_{head}(h(w_i) | w_i, S )
\end{equation}
where~$\mathcal{T}$ is the training set, $h(w_i)$~is $w_i$'s~gold
standard head\footnote{Note that $h(w_i)$ can be {\sc root}.} within
sentence~$S$, and $N_S$~the number of words in~$S$ (excluding {\sc
	root}). During inference, for each word $w_i~(i \in [1, N_S])$ in
$S$, we greedily choose the most likely head $w_j~(j \in [0, N_S])$:
\begin{equation}
	\label{eq:inference}
	w_j = \argmax_{w_j: j \in [0, N_S]} P_{head}(w_j|w_i, S)
\end{equation}
Note that the prediction for each word~$w_i$ is made independently of
the other words in the sentence.

Given our greedy inference method, there is no guarantee that
predicted $\langle$head, dependent$\rangle$ arcs form a tree (maybe
there are cycles). However, we empirically observed that most outputs
during inference are indeed trees. For instance, on an English
dataset, 95\% of the arcs predicted on the development set are trees,
and 87\%~of them are projective, whereas on a Chinese dataset, 87\% of
the arcs form trees, 73\% of which are projective. This indicates that
although the model does not explicitly model tree structure during
training, it is able to figure out from the data (which consists of
trees) that it should predict them.


So far we have focused on unlabeled dependencies, however it is
relatively straightforward to extend \textsc{DeNSe} to produce labeled
dependencies. We basically train an additional classifier to predict
labels for the arcs which have been already identified. The classifier
takes as input features
$[\mathbf{a}_i; \mathbf{a}_j; \mathbf{x}_i; \mathbf{x}_j]$
representing properties of the arc $\langle w_j, w_i \rangle$. These
consist of~$\mathbf{a}_i$ and $\mathbf{a}_j$, the LSTM-based
representations for $w_i$ and $w_j$ (see Equation~\eqref{eq:ann}), and
their word and part-of-speech embeddings, $\mathbf{x}_i$ and
$\mathbf{x}_j$ (see Equation~\eqref{eq:fearep}).  Specifically, we use
a trained {\sc DeNSe} model to go through the training corpus and
generate features and corresponding dependency labels as training
data. We employ a two-layer rectifier network \cite{glorot:2011} for
the  classification task.


\subsection{Maximum Spanning Tree Algorithms}
\label{sec:maxim-spann-tree}

As mentioned earlier, greedy inference may not produce well-formed
trees. In this case, the output of \textsc{DeNSe} can be adjusted with
a maximum spanning tree algorithm.  We use the Chu-Liu-Edmonds
algorithm \cite{chu1965shortest,edmonds1967optimum} for building
non-projective trees and the Eisner \shortcite{eisner1996three}
algorithm for projective ones.

Following \newcite{mcdonald2005non}, we view a sentence
$S=(w_0=\text{\sc root}, w_1, \dots, w_N)$ as a graph
$G_S=\langle V_S, E_S \rangle$ with the sentence words and the dummy
root symbol as vertices and a directed edge between every pair of
distinct words and from the root symbol to every word. The directed
graph~$G_S$ is defined as:
\begin{subequations}
	\begin{align*}
		V_S &= \{w_0 = \text{\sc root}, w_1, \dots ,w_N\} \\
		E_S &= \{ \langle i, j \rangle: i \ne j, \langle i, j \rangle \in [0, N] \times [1, N] \} \\
		s(i, j) &= P_{head}(w_i | w_j, S) \quad \langle i, j \rangle \in E_S
	\end{align*}
\end{subequations}
where $s(i, j)$ is the weight of edge~$\langle i, j \rangle$
and~$P_{head}(w_i | w_j, S)$ is known. The problem of dependency
parsing now boils down to finding the tree with the highest score
which is equivalent to finding a MST in~$G_S$ \cite{mcdonald2005non}.

\paragraph{Non-projective Parsing}
To build a non-projective parser, we solve the MST problem with the
Chu-Liu-Edmonds algorithm \cite{chu1965shortest,edmonds1967optimum}.
The algorithm selects for each vertex (excluding {\sc root}) the
in-coming edge with the highest weight. If a tree results, it must be
the maximum spanning tree and the algorithm terminates. Otherwise,
there must be a cycle which the algorithm identifies, contracts into a
single vertex and recalculates edge weights going into and out of the
cycle.  The greedy inference strategy described in
Equation~\eqref{eq:inference}) is essentially a sub-procedure in the
Chu-Liu-Edmonds algorithm with the algorithm terminating after the
first iteration. In implementation, we only run the Chu-Liu-Edmonds
algorithm through graphs with cycles, i.e.,~non-tree outputs.


\paragraph{Projective Parsing}
For projective parsing, we solve the MST problem with the Eisner
\shortcite{eisner1996three} algorithm.
The time complexity of the Eisner algorithm is $O(N^3)$, while
checking if a tree is projective can be done reasonably faster, with a
$O(N\log N)$~algorithm.
Therefore, we only apply the Eisner algorithm to the non-projective
output of our greedy inference strategy.  Finally, it should be noted
that the \emph{training} of our model does not rely on the
Chu-Liu-Edmonds or Eisner algorithm, or any other graph-based
algorithm. MST algorithms are only used at \emph{test} time to correct
non-tree outputs which are a minority; {\sc DeNSe} acquires underlying
tree structure constraints from the data without an explicit learning
algorithm.


\section{Experiments}
\label{sec:experiments}
We evaluated our parser in a projective and non-projective setting.
In the following, we describe the datasets we used and provide
training details for our models. We also present comparisons against
multiple previous systems and analyze the parser's output.

\subsection{Datasets}
In the projective setting, we assessed the performance of our parser
on the English Penn Treebank (PTB) and the Chinese Treebank 5.1
(CTB). Our experimental setup closely follows \newcite{chen2014fast}
and \newcite{dyer2015transition}.

For English, we adopted the Stanford basic dependencies (SD)
representation \cite{demarneffe:2006}.\footnote{We obtained SD
  representations using the Stanford parser v.3.3.0.} We follow the
standard splits of PTB, sections~2--21 were used for training,
section~22 for development, and section~23 for testing. POS tags were
assigned using the Stanford tagger \cite{toutanova:2003} with an
accuracy of~97.3\%.  For Chinese, we follow the same split of {\tt
  CTB5} introduced in \newcite{zhang2008tale}. In particular, we used
sections \mbox{001--815}, \mbox{1001--1136} for training, sections
\mbox{886--931}, \mbox{1148--1151} for development, and sections
\mbox{816--885}, \mbox{1137--1147} for testing. The original
constituency trees in CTB were converted to dependency trees with the
Penn2Malt tool.\footnote{
  \url{http://stp.lingfil.uu.se/~nivre/research/Penn2Malt.html}} We
used gold segmentation and gold POS tags as in \newcite{chen2014fast}
and \newcite{dyer2015transition}.

In the non-projective setting, we assessed the performance of our
parser on Czech and German, the largest non-projective datasets
released as part of the CoNLL 2006 multilingual dependency parsing
shared task. Since there is no official development set in either
dataset, we used the last~374/367 sentences in the Czech/German
training set as development data.\footnote{We make the number of
  sentences in the development and test sets comparable.} Projective
statistics of the four datasets are summarized in
Table~\ref{tab:proj}.

\begin{table}[t]
	\centering
	\begin{tabular}{ |l | c | c| }
		\hline
		{Dataset} & \# Sentences  &  (\%)  Projective \\
		\hline
		\hline
		English    & 39,832 & \hspace*{1ex}99.9  \\
		Chinese    & 16,091 & 100.0 \\
		Czech  & 72,319 & \hspace*{1ex}76.9 \\
		German  & 38,845 & \hspace*{1ex}72.2 \\
		\hline
	\end{tabular}
	\caption{Projective statistics on four datasets. Number of
		sentences and percentage of projective trees  are calculated
		on the training set.}
	\label{tab:proj}
\end{table}

\subsection{Training Details}
We trained our models on an Nvidia GPU card; training takes one to two
hours. Model parameters were uniformly initialized to~$[-0.1, 0.1]$.
We used Adam \cite{kingma:2014} to optimize our models with
hyper-parameters recommended by the authors (i.e.,~learning
rate~0.001, first momentum coefficient~0.9, and second momentum
coefficient~0.999). To alleviate the gradient exploding problem, we
rescaled the gradient when its norm exceeded~5
\cite{pascanu:2013}. Dropout \cite{srivastava:2014} was applied to our
model with the strategy recommended in the literature
\cite{zaremba:2014,semeniuta-etal-2016:COLING}. On all datasets, we
used two-layer LSTMs and set $d=s=300$, where $d$ is the hidden unit
size and $s$ is the word embedding size.

As in previous neural dependency parsing work
\cite{chen2014fast,dyer2015transition}, we used pre-trained word
vectors to initialize our word embedding matrix $\mathbf{W}_e$. For
the PTB experiments, we used 300 dimensional pre-trained
GloVe\footnote{\url{http://nlp.stanford.edu/projects/glove/}}
vectors \cite{pennington:2014}. For the CTB experiments, we trained
300 dimensional GloVe vectors on the Chinese Gigaword corpus which we
segmented with the Stanford Chinese Segmenter \cite{tseng:2005}. For
Czech and German, we did not use pre-trained word vectors.  The POS
tag embedding size was set to $q=30$ in the English experiments,
$q=50$ in the Chinese experiments and $q=40$ in both Czech and German
experiments.

\subsection{Results}
\label{sec:projr}

For both English and Chinese experiments, we report unlabeled (UAS)
and labeled attachment scores (LAS) on the development and test sets;
following \newcite{chen2014fast} punctuation is excluded from the
evaluation.


Experimental results on PTB are shown in
Table~\ref{tab:en_depparse}. We compared our model with several recent
papers following the same evaluation protocol and experimental
settings.  The first block in the table contains mostly graph-based
parsers which do not use neural networks: Bohnet10 \cite{bohnet:2010},
Martins13 \cite{martins:2013}, and Z\&M14
\cite{zhang:2014:acl}. Z\&N11 \cite{zhang2011transition} is a
transition-based parser with non-local features.  Accuracy results for
all four parsers are reported in \newcite{weiss:2015}.

\begin{table}[t]
	\centering
	\begin{tabular}{ |@{~}l|cc|cc@{~}| }
		\hline
		& \multicolumn{2}{c|}{Dev} & \multicolumn{2}{c|}{Test} \\
		\multicolumn{1}{|c|}{Parser}& UAS & LAS & UAS & LAS \\
		\hline
		\hline
		Bohnet10 & --- & --- & 92.88 & 90.71 \\
		Martins13 & --- & --- & 92.89 & 90.55 \\
		Z\&M14 & --- & --- & 93.22 & 91.02 \\
		Z\&N11 & --- & --- & 93.00 & 90.95 \\
		\hline
		C\&M14 & 92.00  & 89.70  & 91.80  & 89.60  \\
		Dyer15    & 93.20  & 90.90  & 93.10  & 90.90  \\
		Weiss15 & --- & --- & 93.99 & {92.05} \\
		Andor16 & --- & --- & {\bf 94.61} & {\bf 92.79} \\
		K\&G16 {\it graph} & --- & --- & 93.10 & 91.00 \\
		K\&G16 {\it trans} & --- & --- & 93.90 & 91.90 \\
		\hline
		{\sc DeNSe}-Pei & 90.77 & 88.35  & 90.39 & 88.05 \\
		{\sc DeNSe}-Pei+E & 91.39 & 88.94  & 91.00 & 88.61 \\
		\hline
		{\sc DeNSe} & 94.17 & 91.82  & 94.02 & 91.84 \\
		{\sc DeNSe}+E & {\bf 94.30} & {\bf 91.95} & {94.10} & 91.90 \\
		\hline
	\end{tabular}
	\caption{Results on English dataset (PTB with Stanford
          Dependencies). +E:~we post-process  non-projective
          output with the Eisner algorithm.}
	\label{tab:en_depparse}
\end{table}

The second block in Table~\ref{tab:en_depparse} presents results
obtained from neural network-based parsers.  C\&M14
\cite{chen2014fast} is a transition-based parser using features
learned with a feed forward neural network. Although very fast, its
performance is inferior compared to graph-based parsers or strong
non-neural transition based parsers (e.g.,~Z\&N11). Dyer15
\cite{dyer2015transition} uses (stack) LSTMs to model the states of
the buffer, the stack, and the action sequence of a transition
system. Weiss15 \cite{weiss:2015} is another transition-based parser,
with a more elaborate training procedure. Features are learned with a
neural network model similar to C\&M14, but much larger with two
layers. The hidden states of the neural network are then used to train
a structured perceptron for better beam search decoding. Andor16
\cite{andor-EtAl:2016} is similar to Weiss15, but uses a globally
normalized training algorithm instead.

\begin{table}[t]
	\centering
	\begin{tabular}{ |@{~}l|c@{\hspace{1ex}}c|c@{\hspace{1ex}}c@{~}| }
		\hline
		& \multicolumn{2}{c|}{Dev} & \multicolumn{2}{c|}{Test} \\
		\multicolumn{1}{|c|}{Parser}& UAS & LAS & UAS & LAS \\
		\hline
		\hline
		Z\&N11 & --- & --- & 86.00 & 84.40 \\ 
		Z\&M14 & --- & --- & {\bf 87.96} & {\bf 86.34} \\
		\hline
		C\&M14 & 84.00  & 82.40  & 83.90  & 82.40  \\
		Dyer15    & 87.20  & {\bf 85.90}  & 87.20 & 85.70  \\
		K\&G16 {\it graph} & --- & --- & 86.60 & 85.10 \\
		K\&G16 {\it trans} & --- & --- & 87.60 & 86.10 \\
		\hline
		{\sc DeNSe}-Pei & {82.50} & 80.74  & {82.38} & {80.55}  \\
		{\sc DeNSe}-Pei+E & {83.40} & 81.63  & {83.46} & {81.65}  \\
		\hline
		{\sc DeNSe} & {87.27} & 85.73  & {87.63} & {85.94}  \\
		{\sc DeNSe}+E & {\bf 87.35} & 85.85  & {87.84} & {86.15} \\
		\hline
	\end{tabular}
	\caption{Results on Chinese dataset (CTB). +E:~we
          post-process non-projective outputs with the Eisner
          algorithm.}
	\label{tab:ch_depparse}
\end{table}

\begin{table}[t]
	\centering
	\begin{tabular}{ | l |c c |c c| }
		\hline
		& \multicolumn{2}{c|}{PTB} &\multicolumn{2}{c|}{CTB} \\
		Parser & Dev & Test & Dev & Test \\ 
		\hline
		\hline
		C\&M14  & 43.35 & 40.93 & 32.75 & 32.20  \\
		Dyer15 & 51.94 & 50.70 & {\bf 39.72} & {\bf 37.23} \\
		{\sc DeNSe} & 51.24 & 49.34  & 34.74 & 33.66 \\
		{\sc DeNSe}+E & {\bf 52.47} & {\bf 50.79} &36.49 & 35.13 \\
		\hline
	\end{tabular}
	\caption{UEM results on PTB and CTB.}
	\label{tab:uem}
\end{table}

Unlike all models above, {\sc DeNSe} does not use any kind of
transition- or graph-based algorithm during training and
inference. Nonetheless, it obtains a UAS of~94.02\%. Around 95\% of
the model's outputs after inference are trees, 87\% of which are
projective.
When we post-process the remaining 13\% of non-projective outputs with
the Eisner algorithm ({\sc DeNSe+E}), we obtain a slight improvement on UAS (94.10\%).

\newcite{DBLP:journals/corr/KiperwasserG16a} extract features from
bidirectional LSTMs and feed them to a graph- (K\&G16 {\it graph}) and
transition-based parser (K\&G16 {\it trans}).  Their LSTMs are jointly
trained with the parser objective.  {\sc DeNSe} yields very similar
performance to their transition-based parser while it outperforms
K\&G16 {\it graph}.  A key difference between {\sc DeNSe} and K\&G16
lies in the training objective. The objective of {\sc DeNSe} is 
 log-likelihood based \emph{without} tree structure constraints (the model is
trained to produce a distribution over possible heads for each word,
where each head selection is independent), while K\&G16 employ a
max-margin objective \emph{with} tree structure constraints. 
Although our probabilistic objective is non-structured, it is perhaps easier to train compared to a margin-based one.

We also assessed the importance of the bidirectional LSTM on its own
by replacing our LSTM-based features with those obtained from a
feed-forward network. Specifically, we used the 1-order-atomic
features introduced in \newcite{pei2015effective} which represent
POS-tags, modifiers, heads, and their relative positions. As can be
seen in Table~\ref{tab:en_depparse} ({\sc DeNSe}-Pei), these features
are less effective compared to LSTM-based ones and the contribution of
the MST algorithm (Eisner) during decoding is more pronounced ({\sc
  DeNSe}-Pei+E). We observe similar trends in the Chinese, German, and
Czech datasets (see Tables~\ref{tab:ch_depparse}
and~\ref{tab:ger_depparse}).




\begin{figure*}[t]
\hspace*{-2ex}\begin{tabular}{ll}
a. & b. \\
	 \includegraphics[width=0.49\textwidth]{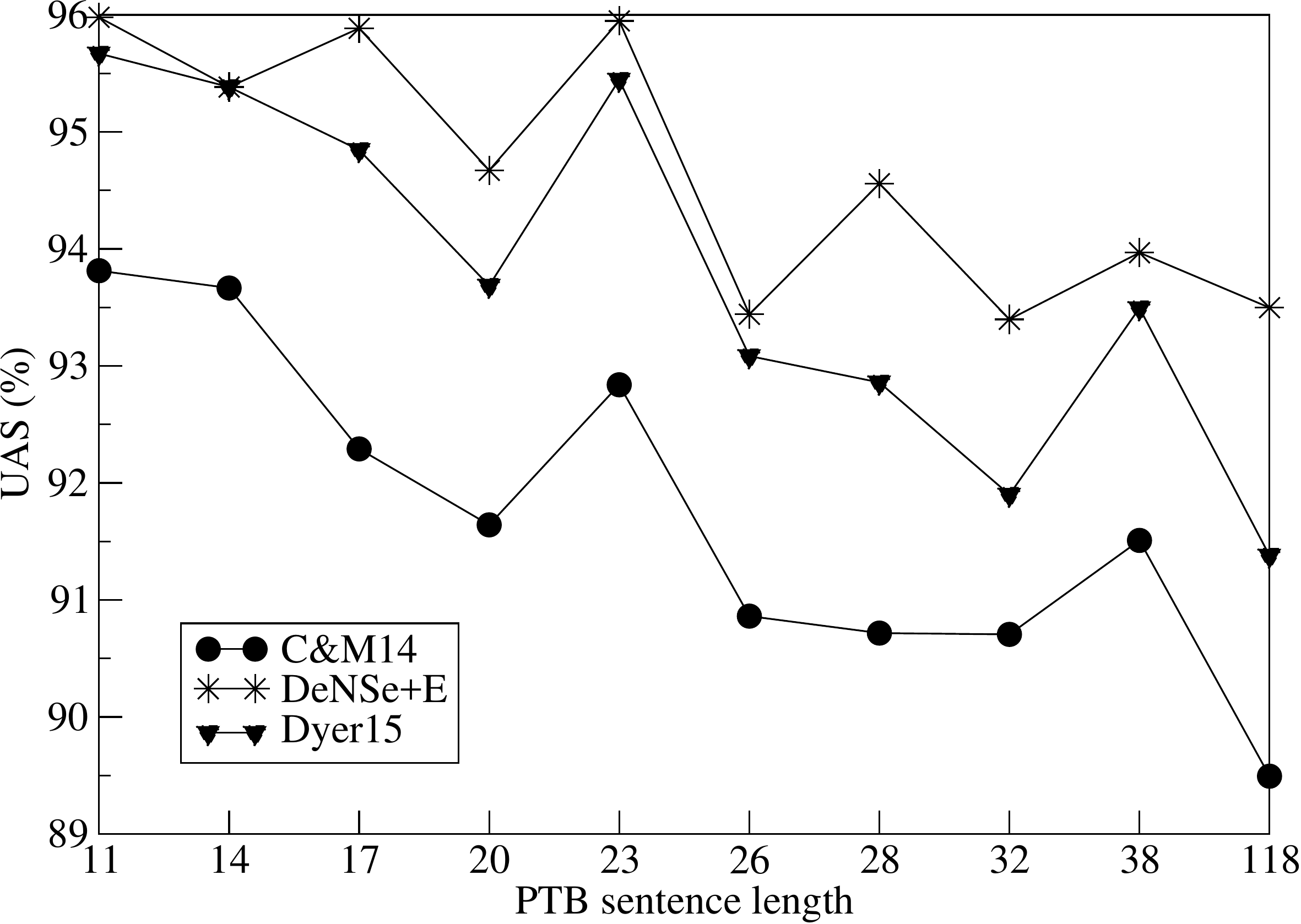}& 
	 \includegraphics[width=0.49\textwidth]{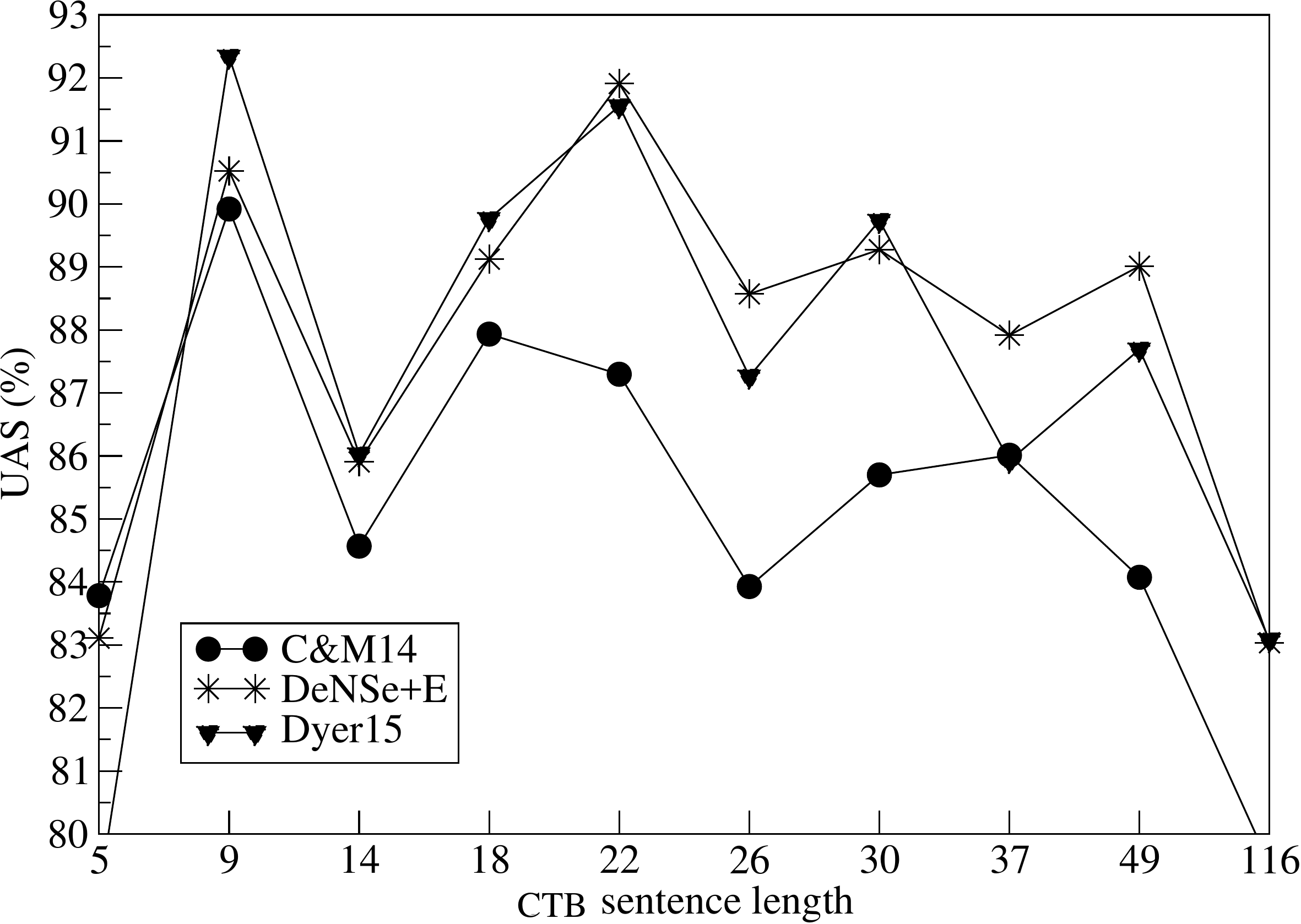}\\
 \\
\end{tabular}
\vspace{-.5cm}
\caption{UAS against sentence length on PTB and CTB (development
  set). Sentences are sorted by length in ascending order and divided
  equally into 10 bins. The horizontal axis is the length of the last
  sentence in each bin.}
	\label{fig:uaslen}
\end{figure*}

\begin{table}[t]
	\centering
	\begin{tabular}{ |@{~}l@{~}|c@{\hspace{1ex}}c@{~}|c@{\hspace{1ex}}c@{~}| }
		\hline
		& \multicolumn{2}{c|}{Czech} & \multicolumn{2}{c|}{German} \\
		\multicolumn{1}{|c|}{Parser} & UAS & LAS & UAS & LAS \\
		\hline
		\hline
		MST-1st & 86.18 & --- & 89.54 & --- \\
		MST-2nd & 87.30 & --- & 90.14  & --- \\
		Turbo-1st & 87.66 & --- & 90.52 & --- \\
		Turbo-3rd & 90.32 & --- & {\bf 92.41} & --- \\
		RBG-1st & 87.90 & --- & 90.24 & ---  \\
		RBG-3rd & {\bf 90.50} & --- & 91.97  &  --- \\
		\hline
		{\sc DeNSe}-Pei & 86.00 & 77.92 & 89.42 & 86.48 \\
		{\sc DeNSe}-Pei+CLE & 86.52 & 78.42 & 89.52 & 86.58 \\
		\hline
		{\sc DeNSe} & 89.60 &  81.70  & 92.15 & 89.58 \\
		{\sc DeNSe}+CLE & 89.68 & 81.72 & 92.19 & 89.60 \\
		\hline
	\end{tabular}
	\caption{Non-projective results on the CoNLL 2006
          dataset. +CLE:~we post-process non-tree outputs
          with the Chu-Liu-Edmonds algorithm.} 
	\label{tab:ger_depparse}
\end{table}

Results on CTB follow a similar pattern. As shown in
Table~\ref{tab:ch_depparse}, \textsc{DeNSe} outperforms all previous
neural models (see the test set columns) on UAS and LAS.
\textsc{DeNSe} performs competitively with Z\&M14, a non-neural model
with a complex high order decoding algorithm involving cube pruning
and strategies for encouraging diversity. Post-processing the output
of the parser with the Eisner algorithm generally improves performance
(by 0.21\%; see last row in Table~\ref{tab:ch_depparse}).  Again we
observe that 1-order-atomic features \cite{pei2015effective} are
inferior compared to the LSTM. Table~\ref{tab:uem} reports unlabeled
sentence level exact match (UEM) in Table~\ref{tab:uem} for English
and Chinese. Interestingly, even when using the greedy inference
strategy, {\sc DeNSe} yields a UEM comparable to Dyer15 on
PTB. Finally, in Figure~\ref{fig:uaslen} we analyze the performance of
our parser on sentences of different length. On both PTB and CTB,
\textsc{DeNSe} has an advantage on long sentences compared to C\&M14
and Dyer15.



For Czech and German, we closely follow the evaluation setup of CoNLL
2006. We report both UAS and LAS, although most previous work has
focused on UAS. Our results are summarized in
Table~\ref{tab:ger_depparse}. We compare {\sc DeNSe} against three
non-projective graph-based dependency parsers: the MST parser
\cite{mcdonald2005non}, the Turbo parser \cite{martins:2013}, and the
RBG parser \cite{lei2014low}. We show the performance of these parsers
in the first order setting (e.g.,~\mbox{MST-1st}) and in higher order
settings (e.g.,~\mbox{Turbo-3rd}). The results of \mbox{MST-1st},
MST-2nd, RBG-1st and RBG-3rd are reported in \newcite{lei2014low} and
the results of Turbo-1st and Turbo-3rd are reported in
\newcite{martins:2013}.  We show results for our parser with greedy
inference (see \textsc{DeNSe} in the table) and when we use the
Chu-Liu-Edmonds algorithm to post-process non-tree outputs ({\sc
  DeNSe}+CLE).

As can been seen, \textsc{DeNSe} outperforms all other first (and
second) order parsers on both German and Czech. As in the projective
experiments, we observe slight a improvement (on both UAS and LAS)
when using a MST algorithm. On German, {\sc DeNSe} is comparable with
the best third-order parser (\mbox{Turbo-3rd}), while on Czech it lags
behind Turbo-3rd and RBG-3rd. This is not surprising considering that
\textsc{DeNSe} is a first-order parser and only uses words and POS
tags as features. Comparison systems use a plethora of hand-crafted
features and more sophisticated high-order decoding
algorithms. Finally, note that a version of \textsc{DeNSe} with
 features in \cite{pei2015effective} is consistently worse (see the second block in
Table~\ref{tab:ger_depparse}).



\begin{table}[t]
	\centering
	\begin{tabular}{ |@{~}l | r |cc|cc@{~}| }
		\hline
		&   & \multicolumn{2}{c|}{ Before MST} &  \multicolumn{2}{c|}{ After MST} \\
		\multicolumn{1}{|c|}{Dataset} & \multicolumn{1}{|c|}{\#Sent}& Tree & Proj & Tree & Proj \\
		\hline
		\hline
		PTB & 1,700 & 95.1 & 86.6 & 100.0 & 100.0 \\
		CTB & 803 & 87.0 & 73.1 & 100.0 & 100.0  \\
		Czech & 374 & 87.7 & 65.5 & 100.0 & \hspace{1ex}72.7 \\
		German & 367 & 96.7 & 67.3 & 100.0 & \hspace{1ex}68.1 \\
		\hline
	\end{tabular}
	\caption{Percentage of trees and projective trees on
		the development set before and after {\sc DeNSe} uses a MST
		algorithm. On PTB and CTB, we use the Eisner algorithm and
		on Czech and German, we use the Chu-Liu-Edmonds algorithm.} 
	\label{tab:treerate}
\end{table}

Our experimental results demonstrate that using a MST algorithm during
inference can slightly improve the model's performance. We further
examined the extent to which the MST algorithm is necessary for
producing dependency trees. Table~\ref{tab:treerate} shows the
percentage of trees before and after the application of the MST
algorithm across the four languages. In the majority of cases {\sc
	DeNSe} outputs trees (ranging from 87.0\% to 96.7\%) and a
significant proportion of them are projective (ranging from 65.5\% to
86.6\%). Therefore, only a small proportion of outputs (14.0\% on
average) need to be post-processed with the Eisner or Chu-Liu-Edmonds
algorithm.

%


%

\section{Conclusions}
\label{sec:conclusions}

In this work we presented {\sc DeNSe}, a neural dependency parser
which we train without a transition system or graph-based
algorithm. Experimental results show that {\sc DeNSe} achieves
competitive performance across four different languages and can
seamlessly transfer from a projective to a non-projective parser
simply by changing the post-processing MST algorithm during inference.
In the future, we plan to increase the coverage of our parser by using
tri-training techniques \cite{li2014ambiguity} and multi-task learning
\cite{luong:2015}.

\paragraph{Acknowledgments}
We would like to thank Adam Lopez and Frank Keller for their valuable
feedback.  We acknowledge the financial support of the European
Research Council (ERC; award number~681760).

\bibliography{eacl2017}
\bibliographystyle{eacl2017}

\end{document}